\documentclass{article}

\usepackage{microtype}
\usepackage{graphicx}
\usepackage{enumerate}
\usepackage{subfigure}
\usepackage{booktabs} % for professional tables

\usepackage{hyperref}

\usepackage[accepted]{icml2021}

\usepackage{amsmath}
\usepackage{amssymb}
\usepackage{mathtools}
\usepackage{amsthm}

\usepackage[capitalize,noabbrev]{cleveref}

\theoremstyle{plain}
\newtheorem{theorem}{Theorem}[section]

\theoremstyle{definition}

\theoremstyle{remark}

\begin{document}

\twocolumn[
% \icmltitle{Submission and Formatting Instructions for \\
%            International Conference on Machine Learning (ICML 2021)}
\icmltitle{Exploring \& Exploiting High-Order Graph Structure for Sparse Knowledge Graph Completion}



\icmlsetsymbol{equal}{*}

\begin{icmlauthorlist}
\icmlauthor{Tao He}{hit}
\icmlauthor{Ming Liu}{hit}
\icmlauthor{Yixin Cao}{smu}
\icmlauthor{Zekun Wang}{hit}
\icmlauthor{Zihao Zheng}{hit}
\icmlauthor{Zheng Chu}{hit}
\icmlauthor{Bing Qin}{hit}
\end{icmlauthorlist}

\icmlaffiliation{hit}{Research Center for Social Computing and Information Retrieval, Harbin Institute of Technology, Heilongjiang, China}
\icmlaffiliation{smu}{Singapore Management University, Singapore}

\icmlcorrespondingauthor{Tao He}{the@ir.hit.edu.cn}
% \icmlcorrespondingauthor{Eee Pppp}{ep@eden.co.uk}

\icmlkeywords{Machine Learning, ICML}

\vskip 0.3in
]



\printAffiliationsAndNotice{} % otherwise use the standard text.

% \begin{abstract}
% This document provides a basic paper template and submission guidelines.
% Abstracts must be a single paragraph, ideally between 4--6 sentences long.
% Gross violations will trigger corrections at the camera-ready phase.
% \end{abstract}

\begin{abstract}
Sparse knowledge graph (KG) scenarios pose a challenge for previous Knowledge Graph Completion (KGC) methods, that is, the completion performance decreases rapidly with the increase of graph sparsity. This problem is also exacerbated because of the widespread existence of sparse KGs in practical applications. To alleviate this challenge, we present a novel framework, LR-GCN, that is able to automatically capture valuable long-range dependency among entities to supplement insufficient structure features and distill logical reasoning knowledge for sparse KGC. The proposed approach comprises two main components: a GNN-based predictor and a reasoning path distiller. The reasoning path distiller explores high-order graph structures such as reasoning paths and encodes them as rich-semantic edges, explicitly compositing long-range dependencies into the predictor. This step also plays an essential role in densifying KGs, effectively alleviating the sparse issue.
Furthermore, the path distiller further distills logical reasoning knowledge from these mined reasoning paths into the predictor. These two components are jointly optimized using a well-designed variational EM algorithm. Extensive experiments and analyses on four sparse benchmarks demonstrate the effectiveness of our proposed method.
\end{abstract}

\section{Introduction}
\begin{figure}[t]
    \centering
    \includegraphics[width=\linewidth]{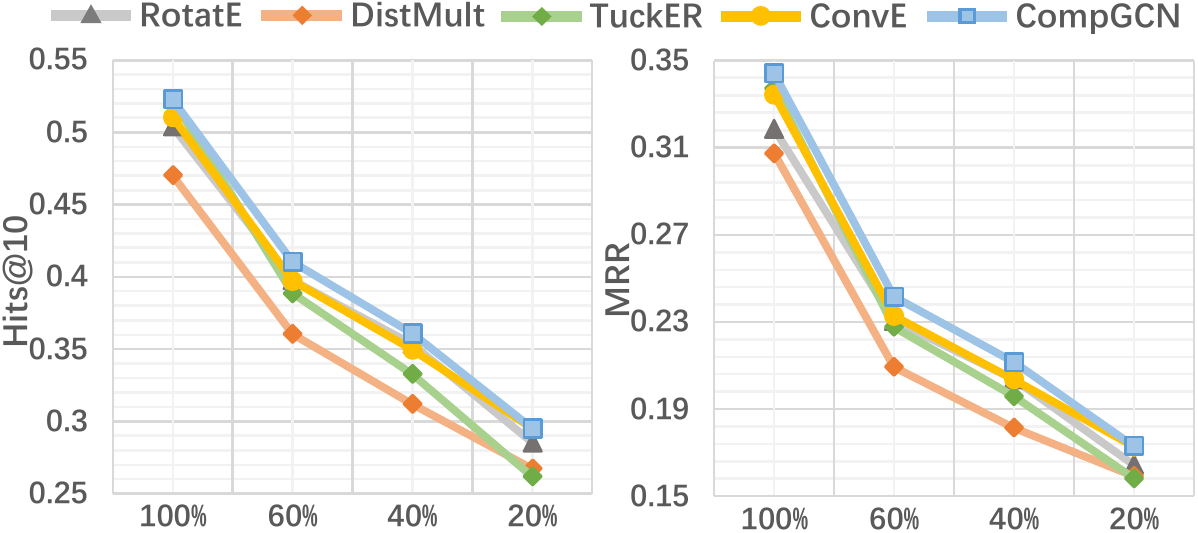}
    \caption{KGC results of different previous KG embedding models on FB15K-237 and its sparse subsets (60\%, 40\%, and 20\% denote percentages of retained triples). The performance drops dramatically as we remove triples.}
    \label{fig:decrease_trend}
\end{figure}

Knowledge Graph Completion (KGC) is the task of reasoning missing facts in the triple format $(h,r,t)$ according to existing facts in a given Knowledge Graph (KG). Despite previous successes on it, KGC often encounters sparsity issues~\cite{lv2020dynamic}. To illustrate this, we conducted pilot studies using five typical KGC models on four datasets with varying sparsity levels. As illustrated in Fig.~\ref{fig:decrease_trend}, the performance curves exhibit a clear downward trend as sparsity increases. Unfortunately, KGs in practical applications are typically much sparser than those in current research ~\cite{chen2022explainable}, considering the insufficient corpus and imperfect effect for the information extraction technology~\cite{Xu2023HowTU,sui2023joint}. Therefore, investigating sparse KGs would greatly benefit real-world applications such as question answering \cite{cao2022kqa,galkin2022inductive}, conversation \cite{li2022c3kg}, and question generation\cite{fei2022lfkqg}.

Previous studies have utilized the graph attention mechanism within the Graph Neural Network (GNN) framework~\cite{chen2022explainable} or contrastive learning~\cite{tan2023kracl} to address the issue of graph sparsity. However, these methods only consider the first-order neighbor structure and disregard rich higher-order graph structure features such as rules and motifs~\cite{jin2022graph}, which are essential semantic components of the graph structure~\cite{yang2018node}. 
A naive solution is to stack K-floor graph convolutional layers~\cite{Kipf2016SemiSupervisedCW} to enable models to consider distant neighbors. However, this approach is susceptible to the over-squashing curse~\cite{Topping2021UnderstandingOA}. This problem states the phenomenon that features from K-hop neighbors are seriously compressed during aggregation processes due to the exponential growth in the number of K-hop neighbors with increasing distances.

% To alleviate this problem, we design an approach that leverages the rich high-order graph structure features by effectively capturing long-range dependencies among entities. 
To this end, we endeavor to design a more effective approach that leverages the rich high-order graph structure features to do the sparse KGC task, where two primary challenges need to be addressed: \textbf{1) Exploration}. Abundant noisy structures are included in exponentially increasing K-hop neighbors. Thus, it is crucial to identify meaningful long-range dependencies while filtering out irrelevant high-order structures. \textbf{2) Exploitation}. GNN methods prioritize the modeling of neighbor correlation, while high-order structure knowledge from motifs or logical rules serves more robust and general. Thus, it is beneficial to integrate high-order structural knowledge into GNN architectures under the sparse KGs.

% To address the issues, we propose a novel framework, LR-GCN, which improves GNN-based methods by capturing reasoning paths for high-order information. 
% LR-GCN consists of two jointly optimized modules, GNN-based predictor and reasoning path Distiller, namely MLN-RL. The basic idea is to drive MLN-RL to discover meaningful reasoning paths via reinforcement learning~\cite{lin2018multi}. Based on that, we further utilize Markov logic Network~\cite{richardson2006markov} (MLN) to distill rule knowledge into GNN-based predictor. Different from neural networks, MLN is a probabilistic logic model that applies Markov Network to first-order logic and enables uncertain inference. We thus can capture true long-range dependency beyond data correlation.

To tackle these issues, we propose a novel framework, LR-GCN, which enhances GNN-based methods by capturing reasoning paths for high-order structure information.
LR-GCN comprises two jointly optimized modules, namely the GNN-based predictor and reasoning path distiller, referred to as MLN-RL. The fundamental motivation is to drive MLN-RL to explore meaningful reasoning paths via Reinforcement Learning~\cite{lin2018multi}, serving as meaningful high-order structures. Subsequently, high-order structure knowledge within these meaningful reasoning paths is exploited both implicitly and explicitly.
% during which two modules are optimized jointly.
Specifically, we propose two joint learning strategies to further leverage high-order structure knowledge among reasoning paths. Firstly, we explicitly incorporate long-range dependencies into graph convolutional layers of the GNN-based predictor by encoding discovered paths into rich-semantic edges to connect distant but relevant neighbors. By doing so, this approach also densifies KGs and alleviates the sparsity issue. Secondly, we implicitly distill logical reasoning abilities within induced rules into the predictor via Markov Logic Network (MLN)~\cite{richardson2006markov}, which is a probabilistic logic model that applies Markov Network to first-order logic and enables uncertain inference. The Variational EM algorithm~\cite{bishop2006pattern} is leveraged to unify the optimization of MLN and GNN-based predictor. 
This strategy constructs unobserved triples as hidden variables to distill first-order rule knowledge into the GNN-based predictor. 
% in the training set based on reasoning paths 
During this process, we expect that the logical reasoning capability of the predictor can be enhanced.
% owing to the generalization ability of logical rules induced from reasoning paths. 
By simultaneously learning these two strategies, we succeed in integrating high-order structure knowledge into embedding learning effectively. 
% Our extensive empirical evaluation shows that LR-GCN achieves prominent relative boosts compared to the backbone model (CompGCN) on four sparse datasets, including FB15K-237\_10 (+16.87\%), FB15K-237\_20 (+9.65\%), NELL23K (+17.04\%) and WD-singer (+5.50\%) at MRR.
Our extensive empirical evaluation shows that LR-GCN boosts the performance of the backbone model (CompGCN) with an obvious margin on four sparse datasets, including FB15K-237\_10 (+4.26\%), FB15K-237\_20 (+2,21\%), NELL23K (+4.37\%) and WD-singer (+2.36\%) at Hits@10.

Our contributions can be summarized as follows: 
\begin{itemize}
    \item We introduce a novel GNN-based framework, LR-GCN, which explores and exploits high-order graph structures to relieve the challenge of sparse KGC.
    \item We propose a novel path-based method, MLN-RL, which generates reasoning paths with delicately calibrated rule weights. This approach effectively filters out noisy paths and explores more instructive high-order graph structures.
    \item We also propose two strategies to explicitly and implicitly exploit the mined high-order graph structure information, including integrating long-range dependencies into the graph convolution and distilling logical reasoning knowledge based on the variational EM algorithm.
    \item Extensive experiments and detailed analyses on four sparse benchmarks consistently demonstrate the effectiveness of our proposed method.
\end{itemize}

% \begin{itemize}
%     \item We are the first to propose a novel GNN-based model that can capture long-range dependencies to relieve the sparse KGC challenge.
%     \item We also first combine the GNN-based model and RL-based method to improve embedding learning.
%     \item We propose a novel RL-based framework MLN-RL that is capable of ranking by answering posterior probabilities and also outputting predictive probabilities.
%     \item Through extensive experiments and detailed analyses on five sparse benchmarks, we empirically demonstrate the effectiveness of our proposed method.
% \end{itemize}

\section{Preliminary}
In this section, we will provide a formalized definition of the knowledge graph completion (KGC) task, along with a set of notations that will be utilized in the framework chapter. Subsequently, we present a concise overview of the utilization of Graph Neural Network models and Reinforcement Learning algorithms in addressing the KGC task, which will facilitate comprehension of the subsequent methods section.

\subsection{Problem Definition}
% (todo: preliminary. including background, e.g., RL and outputed true and false paths...)
A KG is represented by $\mathcal{G}=(\mathcal{E},\mathcal{R},\mathcal{T})$, where $\mathcal{E}=\{e_1,...,e_{|\mathcal{E}|}\}$, $\mathcal{R}=\{r_1,...,r_{|\mathcal{R}|}\}$ denote the entity and relation set, $\mathcal{T}=\{t_i=(e_s,r_q,e_o)\}_{i=1}^{|\mathcal{T}|}$ is the set of observed triples within $\mathcal{G}$. Based on the above definition, KGC task aims to predict $e_o$ given a new query $(e_s,r_q,?)$ or $e_s$ given $(?,r_q,e_o)$. In order to keep uniformity, we follow previous works and refer to both directions as queries with $(e_s,r_q,?)$ by augmenting a reverse relation $r_q^{-1}$ for the query $(?,r_q,e_o)$. Therefore, KGC can be defined as predicting the tail entity $e_o$ for the query $(e_s,r_q,?)$.
%, and without loss of generality, we denote both directions as query $(h,r,?)$ by adding reverse relations for query $(?,r,t)$. 

\subsection{GNN-based KG Embedding}
Our focus is on enhancing embedding learning based on Graph Neural Network~\cite{Kipf2016SemiSupervisedCW} for sparse KGC. In the GNN-based method, given a training case $(e_s,r_q,e_o)$, the KG structure is first encoded with graph convolutional layer(s), resulting in entity and relation embeddings. On top of encoded embeddings, KGE methods such as TransE, DitMult, and ConvE are leveraged to predict the probability distribution over each entity. The aforementioned processes can be formulated as follows:

\begin{equation}
    \mathbf{v_{pred}}=f_{\phi}(e_s,r_q|\mathcal{E}, \mathcal{R})
\end{equation}
where $\mathbf{v_{pred}}\in R^{|\mathcal{E}|}$ denotes the probability vector over all entities, and $\phi$ is the parameters of the model. Finally, the GNN-based model is optimized by BCE Loss as follows:
\begin{equation}
    \mathcal{L}_{label} = \mathbf{BCELoss}(\mathbf{v_{pred}}, e_o)
    \label{eq:label_loss}
\end{equation}

In this paper, we apply CompGCN(-ConvE)~\cite{Vashishth2019CompositionbasedMG}\footnote{Source code is available from https://github.com/malllabiisc/CompGCN.git} as our base GNN-based model. Of course, our framework is agnostic to the particular choice of the GNN-based predictor. Of course, other relevant works such as R-GCN, SACN, and SE-GNN can also be employed in our framework.

\subsection{Reinforcement Learning-based Method}\label{rl-based_method}
\begin{figure}[t]
    \centering
    \includegraphics[width=\linewidth]{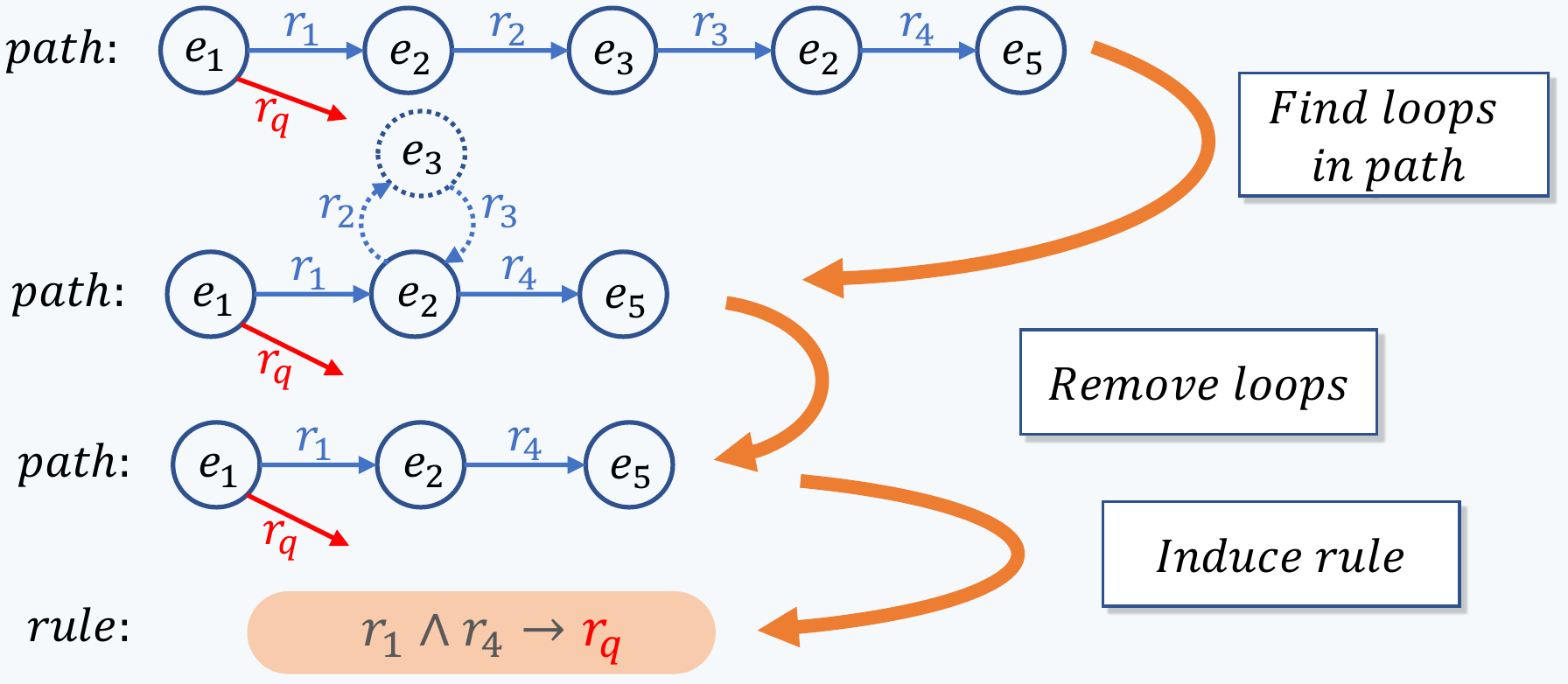}
    \caption{An illustration of inducing rules from reasoning paths. To reduce the length of rules, we view loops within paths as pointless segments and remove them.}
    \label{fig:path_induce_rule}
\end{figure}

The Reinforcement Learning-based (RL-based) approach represents a promising avenue for addressing Knowledge Graph Completion (KGC) by utilizing reinforcement learning to identify interpretable reasoning paths. However, this method often lags behind Knowledge Graph Embedding (KGE) techniques in terms of performance~\cite{lin2018multi,lv2020dynamic}. Specifically, given a query $\mathcal{Q}=(e_s,r_q,?)$, the RL-based method searches for multiple interpretable reasoning paths, represented as $g:(e_s,r_1,e_2)\wedge(e_2,r_2,e_3)\wedge...\wedge(e_{n-1},r_{n-1},e_n)$, and predicts the answer triple $t_c=(e_s,r_q,e_n)$ for each path. The path-searching process is modeled as a Markov Decision Process, with the model expected to induce the inherent rule $rule[g]$ of the form $l:r_1\wedge r_2\wedge ...\wedge r_{n-1}\rightarrow r_q$, as illustrated in Fig.~\ref{fig:path_induce_rule}, where the first part of the rule is called the rule body $l_{[b]}$ and the second part is called the rule head $l_{[h]}$. Moreover, we assign a weight $0\le w_l=p(l_{[h]}|l_{[b]})\le 1$ to each rule $l$ to measure the confidence of the final predictive result. In this study, entities that are not directly connected to the query entity $e_s$ in the path $g$, i.e., $e_3,e_4,...,e_n$, are referred to as high-order neighbors.

In the following section, we base MultihopKG~\cite{lin2018multi}\footnote{Source code is available from https://github.com/salesforce/MultiHopKG.git} to discover helpful paths within sparse KGs.
% In the following section, we base DacKGR~\cite{lv2020dynamic}\footnote{Source code is available from https://github.com/THU-KEG/DacKGR.git} to discover helpful paths since its remarkable improvements over MultihopKG~\cite{lin2018multi} against sparse KGs.
% \subsection{Markov Logic Network}

\begin{figure*}[t]
    \centering
    \includegraphics[width=\linewidth]{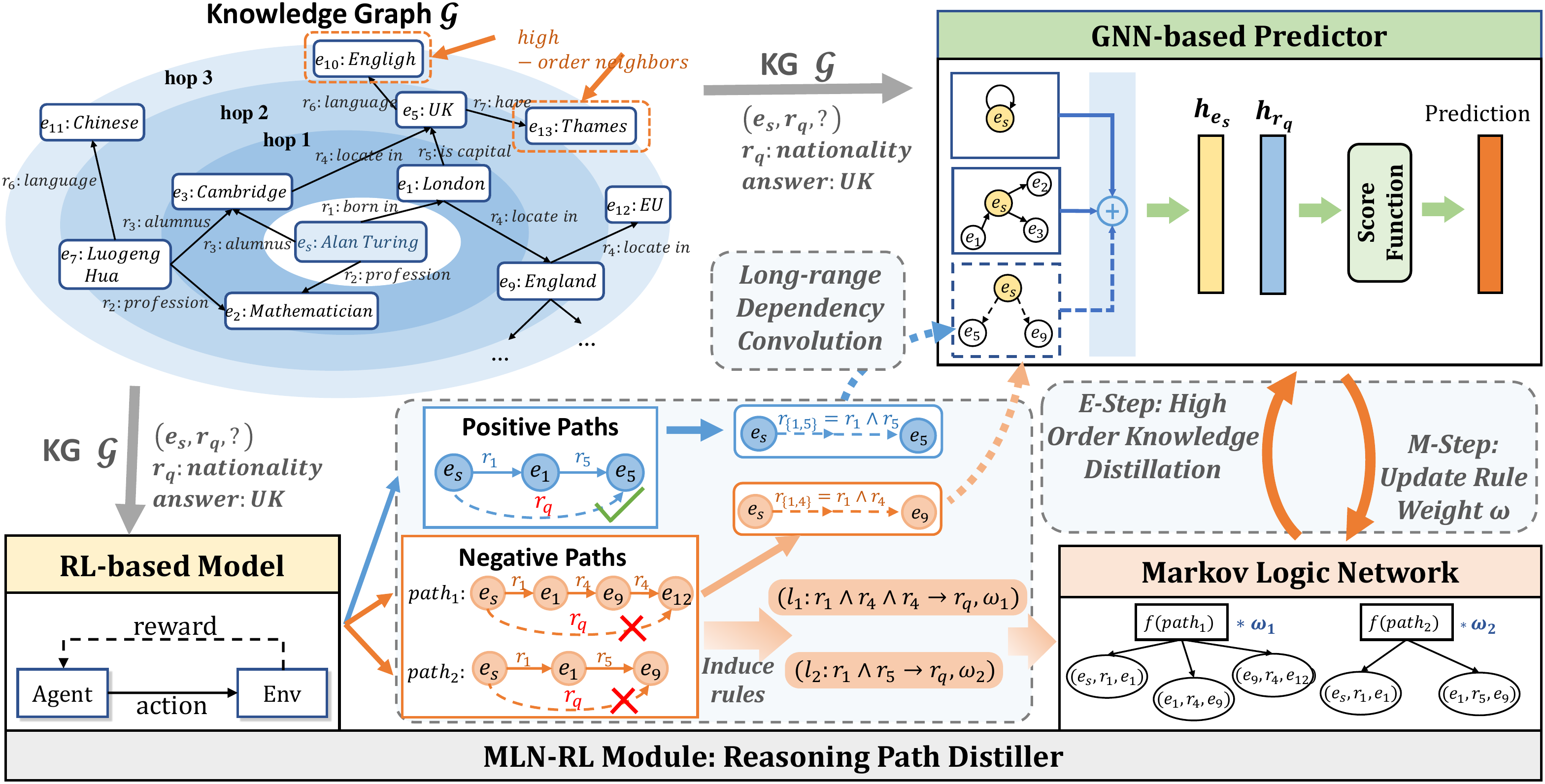}
    \caption{Our framework consists of two modules: GNN-based model with the long range dependency convolution layer and MLN-RL model, which are jointly optimized by high-order knowledge distillation. Given a query $(e_s,r_q,?)$, MLN-RL first reasons paths and classify the paths into two parts according to whether the path is correct. The positive paths and negative path segments are applied to construct new edges to capture long range dependency explicitly. While the negative paths are fed into MLN to distill knowledge for the GNN-based model by variational EM algorithm.} 
    % (todo: 1. colors are not that powerful. 2. not highlight what is high-order structures. 3. ensure the clarity of high-level idea and each component, then add details. )}
    \label{fig:architecture}
\end{figure*}

\section{Framework}
% (todo: highlight relationship among components.)
In this section, we introduce our proposed framework, LR-GCN, for addressing the challenge of sparse Knowledge Graph Completion (KGC), as illustrated in Fig.~\ref{fig:architecture}. Our approach comprises two primary components: the reasoning path distiller module, MLN-RL, and the GNN-based predictor. The MLN-RL module is responsible for exploring reasoning paths that are closely associated with the queries and distilling logical rule knowledge into the GNN-based predictor. The GNN-based model utilizes these reasoning paths as high-order graph structure features to supplement the insufficient 1-hop structural information. Moreover, the MLN-RL modular is able to update weights according to feedback from the GNN predictor, improving the quality of explored reasoning paths and the distilled logical rule knowledge, which in turn enhances the performance of the GNN-based predictor.

To fully leverage the high-order structure knowledge obtained from MLN-RL, we propose two distinct exploitation methods based on the correctness of the reasoning paths. For \textbf{positive paths}, i.e., paths that predict correct entities and all path fragments starting from query entities, we explicitly encode the high-order structure information as long-range dependency knowledge and integrate it into the graph convolutional layers. For \textbf{negative paths}, i.e., paths that predict false answers, rather than simply discarding them, we encourage the GNN-based model to implicitly distill high-order structure knowledge and reasoning capabilities from MLN-RL, as theoretically proven by the variational EM algorithm. These two strategies are optimized upon different categories of reasoning paths without mutual conflicts. The theoretical framework incorporates both strategies, and our empirical experiments demonstrate their cumulative promotions. This is expounded upon in the subsequent experimental section.

In the following sections, we present our proposed reasoning path distiller MLN-RL in section~\ref{Reasoning Path Extractor}. We then elucidate the strategy of utilizing positive reasoning paths within the GNN architecture, which results in the convolution layer structure that accommodates long-range dependencies, as outlined in section~\ref{long-range}. We further expound on the joint learning method between MLN-RL and the GNN-based predictor with respect to negative paths, as detailed in section~\ref{High-order Knowledge Distillation}. This learning approach allows the high-order structure knowledge to be distilled into the GNN-based predictor. Finally, we summarize the total optimization process in section~\ref{optimization_evaluation}.

% view triples concluded by false-prediction paths as hidden variables and formulate the joint distribution of all triplets within paths with a Markov Logic Network (MLN), which is trained with the variational EM algorithm. In E-step learning, we employ the GNN-based model to infer the hidden variables, during which the knowledge induced by the RL agent can be distilled into the GNN-based model. In the M-step, we update the weights of reasoning paths offered by the MLN-RL model to better approximate the posterior probability of predicted entities.

% (the following sections correspond to each component in the fig.3)

% \subsection{GNN-based Predictor}
% \noindent\textbf{Long-dependce xxxx} (todo, description or discussion)
% \subsection{Modules}
% \subsection{Reasoning Path Extractor} \label{Reasoning Path Extractor}
\subsection{MLN-RL: Reasoning Path Distiller} \label{Reasoning Path Extractor}
Current RL-based approaches are limited in only considering the prior weight of the reasoning path $g:(e_s,r_1,e_2)\wedge...\wedge(e_{n-1},r_{n-1},e_n)$, denoted as $w_\theta(g|Q)=p((r_1,e_2)|e_s)p((r_2,e_3)|e_2)...p((r_{n-1}, e_n)|e_{n-1})$, to predict the final answer $e_n$. However, we argue that these approaches neglect the importance of considering the likelihood weight of the reasoning path $g$, denoted as $w_\theta((e_s,r_q,e_n)|g)$. In this study, we introduce an alternative for likelihood weight as $w_{l}=p_w(l_{[h]}|l_{[b]})$, i.e. the confidence of induced rule $l$ from path $g$, where $l_{[h]}$ and $l_{[b]}$ are defined in section~\ref{rl-based_method}. 
Specifically, previous studies have uniformly assigned a value of 1 to the likelihood weight $w_\theta((e_s,r_q,e_n)|g)$. 
Therefore, we define the predictive probability of $e_n$ as follows:
\begin{equation}
    \begin{aligned}
        p_{w,\theta}(e_n|g, \mathcal{Q})&=p_{w,\theta}(t_c=(e_s,r_q,e_n)\text{ is valid}|g,\mathcal{Q})\\
        % &=\sigma(\sum_{g\in \mathcal{P}} w_{rule[g]} w_\theta(e_n|g,\mathcal{Q}))
        &=\sigma(w_{l=rule[g]}\cdot w_\theta(g|\mathcal{Q}))
    \end{aligned}
    \label{eq:likelihood}
\end{equation}
% $\mathcal{P}$ denotes the path set in which the answer entity is $e_n$, and $rule[g]$ refers to the inherent rule of the reasoning path $g$.
where $\mathcal{Q}$ is the input query $(e_s,r_q,?)$. $\sigma$ denotes Sigmoid function and $\theta$ is the parameters of the RL model.

We have also observed that our method bears resemblance to MLN \cite{richardson2006markov} in form. Conventional MLN involves a pre-defined rule set and then the search for all possible rule-induced paths, which can be a time-intensive process. Our approach improves upon this by employing an RL-based technique to first identify paths, followed by the induction of dynamic rules from these paths, which results in a more efficient and real-time updated rule base. As our method is distinct from conventional MLN, we refer to it as MLN-RL. The variational EM algorithm is utilized to update the weights of rules. A more detailed account of our method will be provided in Section~\ref{High-order Knowledge Distillation}.

% \subsection{Joint Learning}
\subsection{GNN-based Predictor with Long Range Dependency Convolution} \label{long-range}
% \noindent\textbf{Long Range Dependency Convolution}
Given a set of positive reasoning paths formulated as $(e_s,r_1,e_2)\wedge(e_2,r_2,e_3)\wedge...\wedge(e_{n-1},r_{n-1},e_n)$ for the query $(e_s,r_q,?)$, 
% we propose the hypothesis that the relation paths abstracted from reasoning paths are similar to the query relation semantically, which can be explained as $r_1\wedge r_2\wedge...\wedge r_{n-1}$ being similar to $r_q$ for the above example. Based on this assumption, 
we first connect $e_s$ with $e_n$ by constructing one new virtual fact $(e_s,r_{1:n-1},e_n)$, where a composite relation $r_{1:n-1}=r_1\wedge r_2\wedge...\wedge r_{n-1}$ is introduced here.
% \footnote{Note that new relations would not bring extra complexity, as these embeddings are obtained by compositing existing relation embeddings and are only used for joint optimization.}.
% Apart from these conclusion facts, we also observe that there are quite a few reasoning paths are false-prediction but part of the reasoning processes are correct. 
We also hope to exploit information from other high-order neighbors for all reasoning paths, even if the path is negative.
For example, for a query $(\textit{Alan Turing}, \textit{language}, ?)$, the reasoning path $(\textit{Alan Turing}, \textit{born in}, \textit{London})\wedge(\textit{London}, \textit{located in}, \textit{UK})\wedge (\textit{UK}, \textit{located in}, \textit{EU})$ is obviously wrong. However, we are still expected to mine helpful information from the path segment $(\textit{Alan Turing}, \textit{born in}, \textit{London})\wedge(\textit{London}, \textit{located in}, \textit{UK})$. To this end, we construct $n-3$ virtual triples $\{(e_s,r_{1:i},e_{i+1})\}_{i=2}^{n-2}$ for each path $(e_s,r_1,e_2)\wedge(e_2,r_2,e_3)\wedge...\wedge(e_{n-1},r_{n-1},e_n)$ by importing $n-3$ composite relations $r_{1:2},r_{1:3},...,r_{1:n-2}$.
% Following the above thoughts, we need to import new relations to construct new triples. 
Specifically, we calculate the embedding $\mathbf{h}_{r_{1:i}}$ for composite relation $r_{1:i}$ as follows:
\begin{align}
    \alpha_{i,j}=&\frac{\sigma(\mathbf{W}_{attn}[\mathbf{h}_{r_q};\mathbf{h}_{r_j}])}{\sum_{k=1}^{i}\sigma(\mathbf{W}_{attn}[\mathbf{h}_{r_q};\mathbf{h}_{r_k}])} \\
    &\mathbf{h}_{r_{1:i}}=\sum_{j=1}^{i}\alpha_{i,j} \mathbf{h}_{r_j}
\end{align}
where $\alpha_{i,j}$ denotes the attention weight of relation $r_j$ in the relation path $r_1\wedge r_2\wedge...\wedge r_{i}$. $[.;.]$ represents the concatenation operator, $\mathbf{W}_{attn}$ is learnable parameters, and $\sigma(.)$ is the LeakyReLU function. 
% $\mathbf{h}_{r_j}$ is the embedding vector for relation $r_j$. 
% With the attention value $\alpha_{i,j}$ obtained, the embeddings of relations within the relation path are summarized to derive a representation $\mathbf{h}_{r_{1:i}}$ for the new relation $r_{1:i}$. 

% \begin{enumerate}[1)]
%     \item SUM module sums relation embeddings that existed in each path directly. That is:

%     \begin{equation}
%         h_{r_{1:i}}=\sum_{i=1}^{i}h_{r_i}
%     \end{equation}
    
%     \item ATTN module sums relation embeddings within each path with separate attention. That is:
%     \begin{equation}
%         h_{r_{1:i}}=\sum_{i=1}^{i}\alpha_i h_{r_i}
%     \end{equation}
%     where $\alpha_i$ denotes the attention value of relation $r_i$ in the relation path $r_1\wedge r_2\wedge...\wedge r_{n-1}$, computed by:
%     \begin{equation}
%         \alpha_i=softmax(LeakyReLU(W_{attn}[h_{r_q};h_{r_i}]))
%     \end{equation}
%     where $[.;.]$ represents concatenating vectors, and $W_{attn}$ is the parameter that needs to be learned.
%     \item LSTM module encodes the relation path using Recurrent Neural Network:
%     \begin{equation}
%         h_{r_{1:i}}=LSTM((h_{r_1},...,h_{r_i});h_0,c_0)
%     \end{equation}
% \end{enumerate}

Upon compressing reasoning paths into new triples $\mathcal{T}_{den}=\{(e_s,r_{1:i},e_{i+1})\}_{i=2}^{n-1}$, the generated triples are integrated into the graph convolutional filter, improving the model's ability to capture long-range dependencies within KGs. To clarify, it is important to note that we do not introduce newly constructed edges to the graph convolution layer explicitly during the training process, as this would alter the structure of the KGs. Instead, we propose the introduction of an additional training loss to incorporate knowledge related to long-range dependencies:
\begin{equation}
    \mathcal{L}_{den}=\frac{1}{|\mathcal{T}_{den}|}\sum_{(e_s,r_{1:i},e_{i+1})\in \mathcal{T}_{den}}\mathbf{BCE}(f_{\phi}(e_s,r_{1:i}), e_{i+1})
    \label{eq:dense_loss}
\end{equation}
% After training with loss $\mathcal{L}_{den}$, the GNN-based predictor is proposed to capture the long-range dependency connected by encoded paths.

\subsection{High-order Knowledge Distillation} \label{High-order Knowledge Distillation}
Negative reasoning paths are also treasure troves with much to be utilized. For example, some paths may actually be false-negative resulting from the inadequacy of Knowledge Graphs (KGs). In order to effectively make use of these negative paths, we view facts within the negative path $g:(e_s,r_1,e_2)\wedge(e_2,r_2,e_3)\wedge...\wedge(e_{n-1},r_{n-1},e_n)$ as observed variables, and the predicted triple $t_c=(e_s,r_q,e_n)$ as hidden variables. As detailed in Section \ref{Reasoning Path Extractor}, building upon previous literature regarding Markov Logic Network~\cite{qu2019probabilistic}, this study models the joint distribution of both observed and hidden variables using MLN-RL as follows:
\begin{equation}
    \begin{aligned}
        &p_{w,\theta}(t_c=(e_s,r_q,e_n), g)
        =\frac{1}{Z(w)}\exp( w_{rule[g]} w_\theta(g|\mathcal{Q}))
        % &=\frac{1}{Z(w)}\exp(\sum_{g\in \mathcal{P}} w_{rule[g]} w_\theta(e_n|g,\mathcal{Q}))
    \end{aligned}
    \label{eq:joint_prob}
\end{equation}
where $Z(w)$ is the partition function, 
% $\mathcal{P}$ denotes the path set in which the answer entity is $e_n$, 
and $rule[g]$ refers to the inherent rule of the reasoning path $g$. $w_{rule[g]}$ and $w_\theta(g|\mathcal{Q})$ are defined in section~\ref{Reasoning Path Extractor}.

The training procedure of MLN-RL is started with maximizing the log-likelihood of the observed facts within the path $g$, i.e., $\log p_{w,\theta}(g)$. We instead optimize the evidence lower bound ($\mathbf{ELBO}$) of the data log-likelihood to introduce hidden variables as follows:
\begin{equation}
    \begin{aligned}
    \log p_{w,\theta}(g)&\ge \mathbf{ELBO}(p_{w,\theta},p_\phi)\\
    &=E_{p_\phi(t_c)}[\log p_{w, \theta}(t_c,g)-\log p_\phi(t_c)]
    \end{aligned}
    \label{eq:em_elbo}
\end{equation}
where $p_\phi$ denotes the GNN-based model. Here, we define the $p_\phi(t_c)$ as the variational distribution. 
% The equality holds if the variational posterior $p_\phi(t_c)$ equals to the true posterior $p_{w,\theta}(t_c|g)$. 
We use the variational EM algorithm \cite{bishop2006pattern} to optimize $\mathbf{ELBO}$, alternating between a variational E-step and an M-step.

% In the varational E-step, we employ a knowledge graph embedding model to infer the missing triplets, during which the knowledge preserved by the logic rules can be effectively distilled into the learned embeddings. In the M-step, the weights of the logic rules are updated based on both the observed triplets and those inferred by the knowledge graph embedding model. In this way, the knowledge graph embedding model provides extra supervision for weight learning. 

\subsubsection{E-Step: Inference}   \label{e_step_inference}
% \noindent\textbf{E-Step: Inference}
In the E-step, we obtain the expectation via updating $p_\phi$ to maximize $\mathbf{ELBO}$. Take Eq.\eqref{eq:joint_prob} into Eq.\eqref{eq:em_elbo}, $\mathbf{ELBO}$ is reorganized as bellow:
\begin{equation}
    \begin{aligned}
        % \mathbf{ELBO}=&E_{p_\phi(t_c)}[\sum_{g\in \mathcal{P}} w_{rule[g]} w_\theta(e_n|g,\mathcal{Q})] \\ &- \log Z(w)-E_{p_\phi(t_c)}[\log p_\phi(t_c)]
        \mathbf{ELBO}=&E_{p_\phi(t_c)}[w_{rule[g]} w_\theta(e_n|g,\mathcal{Q})]\\
        &- \log Z(w)-E_{p_\phi(t_c)}[\log p_\phi(t_c)]
    \end{aligned}
    \label{eq:ELBO}
\end{equation}
where $w_{rule[g]}$ and RL parameters set $\theta$ are fixed in the E-step and thus the partition function $Z(w)$ can be treated as a constant.
% The second term $-E_{p_\phi(t_c)}[\log p_\phi(t_c)]$ is the entropy of variational posterior distribution $p_\phi(t_c)$. 
Here we introduce a theorem to optimize $\mathbf{ELBO}$ in Eq.~\ref{eq:ELBO}.
\begin{theorem}
Optimize $E_{p_\phi(t_c)}[w_{rule[g]} w_\theta(e_n|g,\mathcal{Q})]$ by gradient descent approximates to optimize $\log p_\phi(t_c)[w_{rule[g]} w_\theta(e_n|g,\mathcal{Q})]$:
% For the mean-field variational distribution, the optimal qθ(yn|xV ) of each node n is given by the following fixed-point condition:
\end{theorem}
%
% the environments 'definition', 'lemma', 'proposition', 'corollary',
% 'remark', and 'example' are defined in the LLNCS documentclass as well.
%
% \begin{proof}
% \begin{equation}
%     \begin{aligned}
%         &\nabla_\phi E_{p_\phi(t_c)}[w_{rule[g]} w_\theta(e_n|g,\mathcal{Q})]\\
%         =&\int \nabla_\phi p_\phi(t_c)[w_{rule[g]}w_\theta(e_n|g,\mathcal{Q})]\\
%         =&\int p_\phi(t_c) \nabla_\phi \log p_\phi(t_c)[w_{rule[g]}w_\theta(e_n|g,\mathcal{Q})]\\
%         =&E_{p_\phi(t_c)}[\nabla_\phi \log p_\phi(t_c)[w_{rule[g]}w_\theta(e_n|g,\mathcal{Q})]]\\
%         \approx&\nabla_\phi \log p_\phi(t_c)[w_{rule[g]}w_\theta(e_n|g,\mathcal{Q})]
%     \end{aligned}
% \end{equation}
% \end{proof}
Please refer to Appendix~\ref{app:proof} for the detailed proof. 
Therefore, we optimize $p_\phi$ by minimizing the following loss in the E-step:
\begin{equation}
    \begin{aligned}
        % \mathcal{L}_{elbo}=&-\log p_\phi(t_c)[\sum_{g\in \mathcal{P}}w_{rule[g]}w_\theta(e_n|g,\mathcal{Q})]\\
        % &+E_{p_\phi(t_c)}[\log p_\phi(t_c)]
        \mathcal{L}_{elbo}=&-\log p_\phi(t_c)[w_{rule[g]}w_\theta(e_n|g,\mathcal{Q})]\\
        &+\lambda * E_{p_\phi}[\log p_\phi(t_c)]
    \end{aligned}
    \label{eq:elbo_loss}
\end{equation}
where $\lambda$ is a hyperparameter to be tuned. 
We explain the practical meanings of this optimization objective. The first term requires $p_\phi$, i.e. the GNN-based model, to maximize the likelihood value of the triple predicted by path $g$ weighted by the path importance $w_{rule[g]}w_\theta(e_n|g,\mathcal{Q})$,
% , that is the likelihood function value is weighted by the path quality. 
during which high-order graph structure knowledge contained in reasoning paths can be distilled into the GNN-based model. On the other hand, the second term acts as an entropy constraint, encouraging $p_\phi$ to retain the knowledge it has learned and not to overtrust in MLN-RL. The GNN-based predictor is expected to trade off between retaining its own knowledge and learning from MLN-RL.

\subsubsection{M-Step: Learning}
% \noindent\textbf{M-Step: Learning}
In the M-step, we will fix the GNN-based model $p_\phi$ and update the weights of rules in MLN-RL induced from reasoning paths by maximizing $\mathbf{ELBO}$. However, it is untractable to optimize $\mathbf{ELBO}$ directly as the partition function $Z(w)$ is no longer a constant. Following existing studies \cite{qu2019probabilistic}, we instead optimize the pseudo-likelihood function:
\begin{equation}
\begin{aligned}
    \mathcal{F}_{PL}(w)&= E_{p_\phi(t_c)}[\log p_{w,\theta}(t_c|\mathbf{MB}(t_c))]\\
    &\approx E_{p_\phi(t_c)}[\log p_{w,\theta}(t_c|g)]
\end{aligned}
\end{equation}
where $\mathbf{MB}(t_c)$ is the Markov Blanket of $t_c$, involving triples that appear together with $t_c$ in the groundings of induced rules. $p_{w,\theta}(t_c|g)$ is defined in Eq.\eqref{eq:likelihood}.
% , i.e. reasoning paths.

For each induced rule $l$ from path $g$ that can conclude the triple $t_c$, we optimize the rule weight $w_l$ by gradient descent, with the derivative:
\begin{equation}
\begin{aligned}
    &\nabla_{w_l}E_{p_\phi(t_c)}[\log p_{w,\theta}(t_c|g)]
    \approx p_\phi(t_c=1)-p_{w,\theta}(t_c=1|g)
\end{aligned}
\label{eq:m_step_update_w}
\end{equation}
where $t_c=1$ means $t_c$ is valid. The proof of this conclusion is given by \cite{qu2019probabilistic}.
% The process of proof is presented in Appendix A.2.
Intuitively, for each triple $t_c$ predicted by the false-prediction path $g$, we apply $p_\phi(t_c = 1)$ as the target for updating the probability $p_{w,\theta}(t_c=1|g)$. In this way, the rule weights are updated to better measure the prior importance of reasoning paths.

\subsubsection{Discussion}
For the implementation of E-Step, we optimize $p_\phi$, i.e. the GNN-based predictor, to maximize the likelihood values of not only the triples $t_c$ predicted by MLN-RL but all triples in the corresponding reasoning path, weighted by the path weight. Therefore, the GNN-based predictor learns both the predicted results and reasoning processes from MLN-RL. 
By doing so, the GNN-based predictor promises both to distill high-order structure knowledge from predicted results and to acquire reasoning capabilities from the reasoning processes of MLN-RL.

\subsection{Optimization and Evaluation} \label{optimization_evaluation}
To speed up the training process, we pretrain the base GNN-based model and MLN-RL in advance until convergence. After that, we start jointly learning the GNN-based predictor and MLN-RL for certain epochs.

During the joint learning, reasoning paths are generated using MLN-RL for a given case $(e_s,r_q,e_o)$ using the query $(e_s,r_q,?)$. Three losses are minimized together to train the GNN-based model:
\begin{equation}
    \mathcal{L}_{gnn} = \mathcal{L}_{label} + \beta * \mathcal{L}_{den} + \gamma * \mathcal{L}_{elbo}
    \label{eq:gnn_loss}
\end{equation}
% where $\beta$ and $\gamma$ are weights of $\mathcal{L}_{den}$ in Eq.~\eqref{eq:dense_loss} and $\mathcal{L}_{elbo}$ in Eq.~\eqref{eq:elbo_loss}, respectively, 
where $\mathcal{L}_{label}$, $\mathcal{L}_{den}$ and $\mathcal{L}_{elbo}$ are defined in Eq.~\eqref{eq:label_loss}, Eq.~\eqref{eq:dense_loss} and Eq.~\eqref{eq:elbo_loss}, respectively, and $\beta$ and $\gamma$ are corresponding loss weights.
and we view them as hyperparameters.
The rule weights in MLN-RL are then updated using the derivative in Eq.~\eqref{eq:m_step_update_w} in M-step. Finally, the learned high-order graph structure knowledge from MLN-RL is expected to be integrated into the GNN-based model resulting in LR-GCN model for evaluation.

\section{Experiments}
\subsection{Experimental Settings} \label{Experimental Settings}
% \subsubsection{Datasets}
\noindent\textbf{Datasets}
% Following DacKGR \cite{lv2020dynamic} and HoGRN \cite{chen2022explainable}, we also conduct experiments on two sparse datasets WD-singer and NELL23K.
% % \footnote{\href{https://github.com/THU-KEG/DacKGR}{https://github.com/THU-KEG/DacKGR}}
% % sampled from NELL \cite{carlson2010toward} and Wikidata \cite{vrandevcic2014wikidata}, respectively.
% To explore the performance of our framework under sparse scenarios, we uniformly sample 10\%, 20\%, 30\%, and 60\% triples from FB15K-237 to construct sparser datasets, abridged as FB15K-237\_10, FB15K-237\_20, FB15K-237\_30, and FB15K-237\_60. 
% Differing from \cite{lv2020dynamic}, we reserve all entities and relations by constraining each entity (relation) to participate in at least one triple fact. 
% The statistical and constructed details of datasets are summarized in Table \ref{tab:dataset_statistics}.
This study follows the experimental setups of DacKGR~\cite{lv2020dynamic} and HoGRN~\cite{chen2022explainable} on two sparse datasets, namely WD-singer and NELL23K. To evaluate the performance of our framework in sparse scenarios, we also uniformly sample 10\%, 20\%, 30\%, and 60\% of the triples from FB15K-237, resulting in sparser datasets denoted as FB15K-237\_10, FB15K-237\_20, FB15K-237\_30, and FB15K-237\_60, respectively. Unlike DacKGR~\cite{lv2020dynamic} and HoGRN~\cite{chen2022explainable}, our methodology ensures that all entities and relations are preserved by enforcing each entity (relation) to participate in at least one triple fact. Table~\ref{tab:dataset_statistics} summarizes the dataset statistics. We provide construction details of datasets in Appendix~\ref{dataset_construction_details}. 
% The present research abstains from conducting experiments on the widely used KGC benchmark known as WN18RR~\cite{dettmers2018convolutional} due to its limited number of relations compared with the above datasets, resulting in a reduced capacity for RL modules to extract high-order structures. Our experimental analyses confirm the marginal enhancements achieved on WN18RR.

\begin{table*}[t]
    \centering
    \caption{Summary statistics of datasets}
    \resizebox{\linewidth}{!}{
    \begin{tabular}{l|ccccccc}
    \toprule
     & WD-singer & NELL23K & FB15K-237\_10 & FB15K-237\_20 & FB15K-237\_30 & FB15K-237\_60 & FB15K-237 \\
    \midrule
    \#Entity & 10282 & 22925 & 14541 & 14541 & 14541 & 14541 & 14541 \\
    \#Relation & 270 & 400 & 237 & 237 & 237 & 237 & 237 \\
    \#Train Set & 16142 & 24321 & 27211 & 54423 & 108846 & 163269 & 272115 \\
    \#Dev Set & 2163 & 4951 & 17535 & 17535 & 17535 & 17535 & 17535 \\
    \#Test Set & 2203 & 4944 & 20466 & 20466 & 20466 & 20466 & 20466 \\
    Avg In-degree & 1.570 & 1.170 & 1.876 & 3.752 & 5.628 & 11.256 & 18.760\\
    \bottomrule
    \end{tabular}
    }
    \label{tab:dataset_statistics}
\end{table*}

% \subsubsection{Baselines}
\noindent\textbf{Baselines}
% we mainly compare LR-GCN with the backbone method, i.e. CompGCN~\cite{vashishth2019composition} in the following experiments, to verify the effectiveness of our proposed method. We focus on whether LR-GCN can enhance the backbone model and the extent to which it can do so. We also compare other embedding-based KGC models, including TransE~\cite{bordes2013translating}, RotatE~\cite{sun2019rotate}, ComplEx~\cite{trouillon2016complex}, TuckER~\cite{balavzevic2019tucker}, ConvE~\cite{dettmers2018convolutional}, and two GNN-based method SCAN~\cite{shang2019end}, RED-GNN~\cite{zhang2022knowledge}. In addition, we present results of the relevant work on sparse KGs, i.e. DacKGR \cite{lv2020dynamic}, HoGRN~\cite{chen2022explainable}, for comparison.
% % \footnote{We do not compare with HoGRN because it does not provide the source codes}.
we mainly compare LR-GCN with the backbone method CompGCN~\cite{Vashishth2019CompositionbasedMG} in the following experiments. The primary objective is to assess whether LR-GCN can improve upon the performance of the backbone model and to what extent. Additionally, other KGC models that are based on embeddings such as TransE~\cite{bordes2013translating}, RotatE~\cite{Sun2018RotatEKG}, ComplEx~\cite{trouillon2016complex}, TuckER~\cite{balavzevic2019tucker}, ConvE~\cite{dettmers2018convolutional}, and GNN-based methods like SCAN~\cite{shang2019end}, NBFNet~\cite{zhu2021neural}
% \footnote{Source code is available from https://github.com/DeepGraphLearning/NBFNet.git}
and RED-GNN~\cite{zhang2022knowledge}
% \footnote{Source code is available from https://github.com/LARS-research/RED-GNN.git}
are evaluated. Furthermore, the findings of relevant studies on sparse KGs, namely DacKGR~\cite{lv2020dynamic} and HoGRN~\cite{zhang2022knowledge}, are also presented for comparative analysis.

% \subsubsection{Hyperparameters}
\noindent\textbf{Hyperparameters}
In our implementation, we set the embedding dimension to 200 for all KGE models.
We first pre-train CompGCN as the backbone and then jointly retrain CompGCN and MLN-RL until convergence under the same learning rate.
For fairness, we also retrain CompGCN independently for the same number of epochs. Learning rates for CompGCN and LR-GCN are set to 0.005 on WD-singer and NELL23K, and 0.001 on FB15K-237 and its sub-datasets. For more hyperparameters please refer to Appendix~\ref{app:hyperparamter_setting}.
% For other hyperparameters please refer to Appendix~\ref{app:hyperparamter_setting}.

% \subsubsection{Evaluation Metrics}
\noindent\textbf{Evaluation Metrics}
Same as previous work, we use ranking metrics to evaluate our framework, i.e. MRR and HITS@k. Besides, we filter out all remaining entities valid for the test query $(h, r, ?)$ from the ranking. The metrics are measured in both tail prediction and head prediction directions. We strictly follow the “RANDOM” protocol proposed by \cite{sun2020re} to evaluate our methods.

\begin{table*}[t]
    \centering
    \caption{Experimental results on FB15K-237\_10, FB15K-237\_20, WD-singer, and NELL23K. The last line records the relative improvements of LR-GCN over CompGCN. Hits@N and MRR values are in percentage. ``KD" denotes High-order Knowledge Distillation and ``LRC" represents Long Range Convolution. The best score is in \textbf{bold} and the second is \underline{underlined}.}
    \resizebox{\linewidth}{!}{
    \begin{tabular}{cccccccccccccccc}
    \toprule
    &&&& \multicolumn{3}{c}{FB15K-237\_{10}} & \multicolumn{3}{c}{FB15K-237\_{20}} & \multicolumn{3}{c}{WD-singer} & \multicolumn{3}{c}{NELL23K}\\
    \cmidrule(lr){5-7}\cmidrule(lr){8-10}\cmidrule(lr){11-13}\cmidrule(lr){14-16}
    &&&& \multicolumn{2}{c}{Hits@N $\uparrow$} && \multicolumn{2}{c}{Hits@N$\uparrow$} && \multicolumn{2}{c}{Hits@N$\uparrow$} && \multicolumn{2}{c}{Hits@N$\uparrow$}\\
    \cmidrule(lr){5-6}\cmidrule(lr){8-9}\cmidrule(lr){11-12}\cmidrule(lr){14-15}
    &&&& @1 & @10 & MRR $\uparrow$ & @1 & @10 & MRR $\uparrow$ & @1 & @10 & MRR $\uparrow$ & @1 & @10 & MRR $\uparrow$\\
    \midrule
    \multicolumn{4}{l}{TransE} & 4.94 & 24.03 & 11.58 & 8.33 & 28.76 & 15.27 & 22.56 & \underline{49.84} & 32.68 & 4.62 & 30.21 & 13.35\\
    \multicolumn{4}{l}{RotatE} & 5.64 & 17.16 & 9.52 & 10.40 & 28.52 & 16.43 & 31.43 & 45.96 & 36.79 & 9.63 & 28.13 & 15.75\\
    % \multicolumn{4}{l}{DistMult} & 9.51 & 22.85 & 13.90 & 10.68 & 26.72 & 15.92 & 25.51 & 39.51 & 30.63 & 12.71 & 33.26 & 19.50\\
    \multicolumn{4}{l}{ComplEx} & 9.61 & 22.77 & 13.92 & 10.69 & 26.17 & 15.81 & 29.87 & 43.53 & 34.94 & 14.21 & 32.57 & 20.26\\
    \multicolumn{4}{l}{TuckER} & 6.51 & 16.44 & 9.89 & 9.73 & 25.91 & 15.08 & 32.12 & 44.83 & 36.87 & 13.75 & 30.35 & 19.36\\
    \multicolumn{4}{l}{ConvE} & 10.56 & 23.90 & 15.69 & 11.38 & 29.51 & 17.27 & 31.46 & 47.37 & 37.22 & 14.44 & 37.55 & 22.73\\
    \multicolumn{4}{l}{SACN} & 9.60 & 22.56 & 13.98 & 11.36 & 28.82 & 17.07 & 28.49 & 43.70 & 34.10 & 14.36 & 35.01 & 21.22\\
    \multicolumn{4}{l}{RED-GNN} & 8.43 & 19.8 & 12.22 & {12.28} & 30.88 & {18.43} & {32.57} & 49.18 & {38.79} & \underline{16.63} & 39.65 & {24.24} \\
    \multicolumn{4}{l}{NBFNet} & \textbf{11.13} & \textbf{27.78} & \textbf{16.64} & \underline{12.45} & \textbf{31.87} & \underline{18.89} & \underline{32.59} & 48.10 & {38.81} & {14.40} & 39.04 & {22.74} \\
    \midrule
    \multicolumn{4}{l}{DacKGR} & 10.21 & 21.34 & 13.91 & 10.99 & 26.15 & 15.87 & 26.49 & 44.30 & 32.66 & 13.00 & 31.63 & 18.99\\
     \multicolumn{4}{l}{HoGRN} & - & - & - & - & - & - & - & 48.80 & \underline{39.07} & - & \underline{39.98} & \underline{24.56}\\
    % \multicolumn{4}{l}{KBGAT} & - & - & - & - & - & - & - & - & - & - & - & -\\
    \midrule
    \multicolumn{4}{l}{\textit{-backbone}}\\
    \multicolumn{4}{l}{CompGCN} & 9.52 & 23.11 & 14.11 & 11.27 & 29.51 & 17.30 & 31.75 & 47.62 & 37.65 & 13.79 & 37.57 & 21.59\\
    \midrule
    \multicolumn{4}{l}{LR-GCN (w/o KD)} & 10.41 & {26.32} & 15.77 & {11.94} & {30.98} & {18.22} & {32.06} & 49.61 & {38.71} & 15.41 & {39.84} & 23.41\\
    % \multicolumn{4}{l}{\textit{–Impr}} & () & () & () & () & () & () & () & () & () & () & () & ()\\
    \multicolumn{4}{l}{LR-GCN (w/o LRC)} & {10.60} & 26.25 & {15.89} & 11.59 & 30.11 & 17.74 & 31.89 & 48.28 & 38.00 & {15.99} & 39.11 & {23.53}\\
    % \multicolumn{4}{l}{\textit{–Impr}} & () & () & () & () & () & () & () & () & () & () & () & ()\\
    \multicolumn{4}{l}{LR-GCN} & \underline{11.07} & \underline{27.37} & \underline{16.49} & \textbf{12.57} & \underline{31.72} & \textbf{18.97} & \textbf{32.80} & \textbf{49.98} & \textbf{39.27} & \textbf{17.31} & \textbf{41.90} & \textbf{25.27}\\
    \multicolumn{4}{l}{\textit{–Rela. Impr.}} & +\textit{16.28\%} & +\textit{18.43\%} & +\textit{16.87\%} & +\textit{11.54\%} & +\textit{7.49\%} & +\textit{9.65\%} & +\textit{3.31\%} & +\textit{4.96\%} & +\textit{5.50\%} & +\textit{25.53\%} & +\textit{11.53\%} & +\textit{17.04\%}\\
    \bottomrule
    \end{tabular}
    }
    \label{tab:main_results}
\end{table*}

\subsection{Main Results}
Table~\ref{tab:main_results} presents the main results of three datasets, showcasing the consistent improvements observed in LR-GCN across all sparse datasets. (1) The MRR metric indicates that LR-GCN outperforms CompGCN with relative improvements of 16.87\%, 9.65\%, 5.50\%, and 17.04\%, respectively. These results validate the effectiveness of our proposed method in capturing long-range dependency to supplement insufficient structure features. (2) LR-GCN exhibits more significant improvements across all benchmarks compared to variants without Knowledge Distillation (KD) and Long Range Convolution (LRC) modules. This is because the high-order structure knowledge captured by KD and LRC is different and complementary, resulting in better performance when combined. (3) Our assessment reveals that, compared to NBFNet, our method yields slightly inferior results on FB15K-237\_10 and FB15K-237\_20, primarily due to the limitations inherent in our backbone model CompGCN. However, we want to note that our objectives revolve around utilizing higher-order structural information to enhance the performance of the backbone in sparse KGs and our framework is model-agnostic. We are confident that substituting NBFNet for CompGCN will enable us to achieve superior metrics.

\subsection{KG Sparsity Analysis} \label{exp:sparsity_analysis}
\begin{figure}[t]
    \centering
    \includegraphics[width=\linewidth]{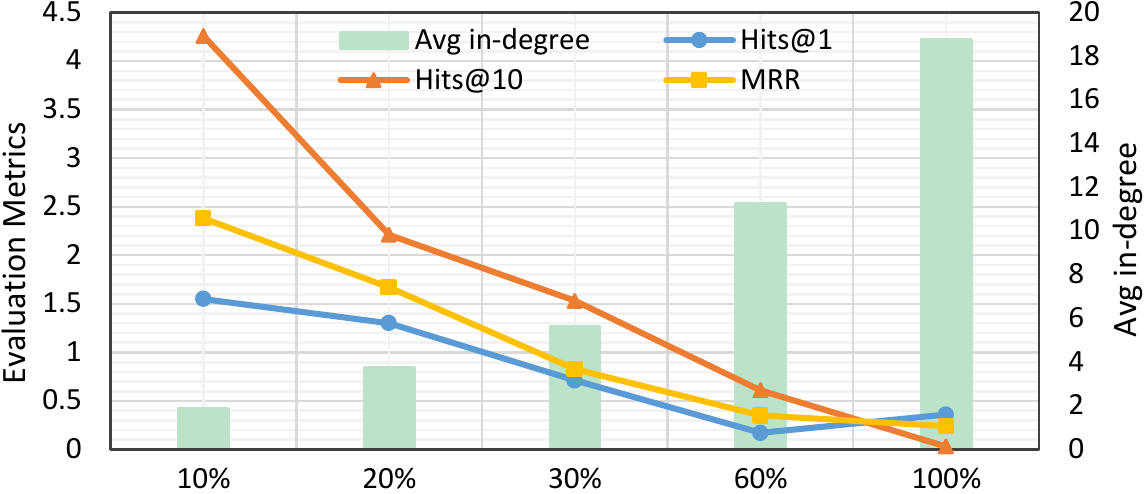}
    \caption{Improvements of LR-GCN on FB15K-237 and 4 sparse datasets against to CompGCN (60\%, 30\%, 20\%, and 10\% denote percentages of retained triples).}
    \label{fig:sparsity_analysis}
\end{figure}

% In this section, we analyze the impacts of KG sparsity based on FB15K-237 and series of sub-datasets FB15K-237\_x (x=10, 20, 30, 60). The results are presented in Figure~\ref{fig:sparsity_analysis}. As we can see, the improvement of LR-GCN against CompGCN increases as the average in-degree decreases, i.e. the graph sparsity degree increases. 
% % LR-GCN improves 2.01/3.57 points at MRR/Hits@10 on FB15K-237\_10 while only achieve 0.62/0.59 points improvements at MRR/Hits@10 on FB15K-237\_30, 
% % which is in line with our expectation, 
% indicating that our method is more superior for sparse KGC. The reason is that entities possess enough first-order neighbors in the dense KGs so that the information gained from higher-order structures is little.
This section presents an analysis of the impact of knowledge graph (KG) sparsity upon FB15K-237 and its sub-datasets FB15K-237\_x (x=10, 20, 30, 60). As explicated in section~\ref{Experimental Settings}, we construct three sub-datasets FB15K-237\_10, FB15K-237\_20, FB15K-237\_30, and FB15K-237\_60 by randomly removing 90\%, 80\%, 70\%, and 40\% triples. With the aim of ensuring equal difficulty levels across tasks with varying graph sparsities, we enforced the constancy of entities and relations among all sub-datasets, thereby guaranteeing the participation of each entity and relation in at least one triplet. The findings, depicted in Fig.~\ref{fig:sparsity_analysis}, reveal that improvements over CompGCN decrease as the average in-degree increases.
% indicating a greater superiority of our method for sparse KGC. 
This is due to the fact that entities in dense KGs possess sufficient first-order neighbors, resulting in limited information gained from higher-order structures.

\subsection{In-degree Analysis} \label{exp:in-degree_analysis}
\begin{figure}[t]
    \centering
    \includegraphics[width=\linewidth]{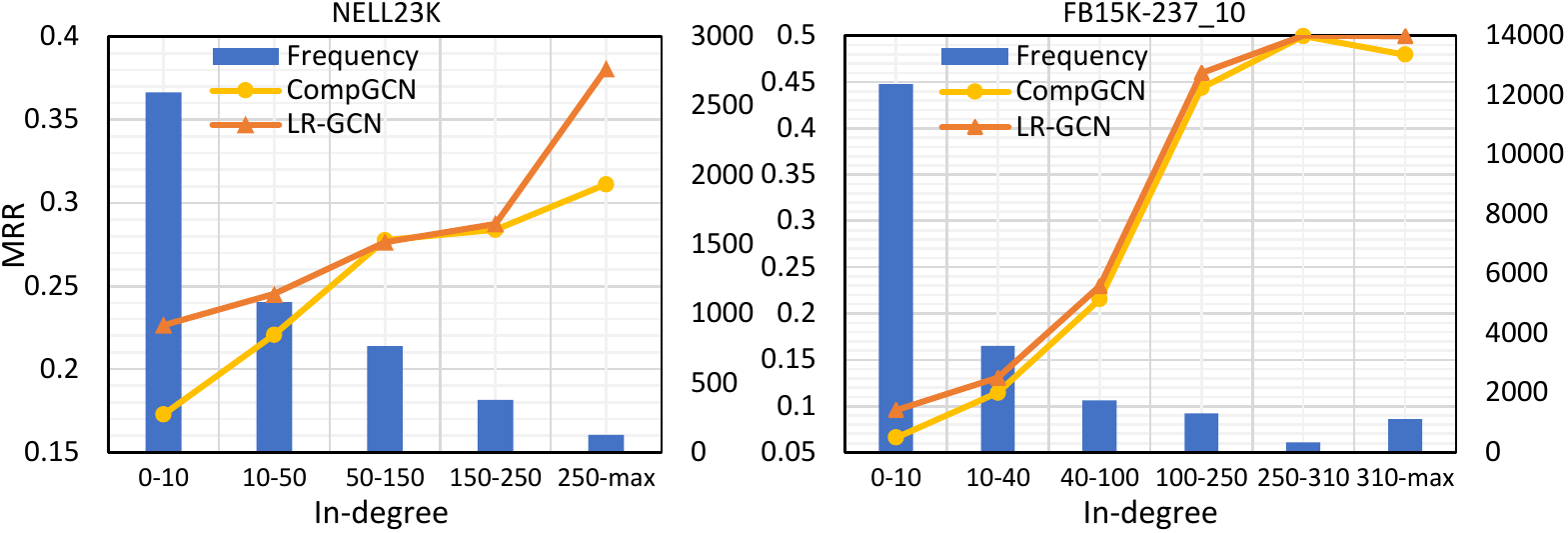}
    \caption{MRR results and entity frequency grouped by entity in-degree on NELL23K and FB15K-237\_10.}
    \label{fig:in-degree_analysis}
\end{figure}

% To better verify LR-GCN's superiority under high-sparsity entities, we follow \cite{shang2019end} and analyze performances on entities with different in-degree. We first group the test set into several subsets according to the in-degree of entities and compare LR-GCN with CompGCN on each subset. From Figure~\ref{fig:in-degree_analysis}, we observe that: (1) Performances of CompGCN and LR-GCN universally decrease as the entities' in-degree decreases, indicating low in-degree, i.e. high sparsity, indeed worsens embedding learning. (2) Improvements from LR-GCN against CompGCN decrease as in-degree increases (except last in-degree scopes), revealing that LR-GCN is more sensitive to high-sparsity entities since high-sparsity entities gain more information from high-order structures. (3) Significant improvements also occur at the maximal in-degree scopes on two datasets. We hypothesize the gains originate from the reasoning capacity distilled from MLN-RL, which is not limited by graph sparsity. More detailed research will be conducted in the future.
This study evaluates the performance of LR-GCN and CompGCN on entities with varying in-degree to determine LR-GCN's superiority under high-sparsity conditions. The test set was divided into subsets based on entity in-degree, and then two models were compared on each subset. Results in Fig.~\ref{fig:in-degree_analysis} showed that: (1) As the in-degree decreases, there is an increase in the entity frequency. This observation suggests that resolving high-sparsity concerns is of great significance in enhancing the efficacy of integral completion, especially for entities with low in-degree. (2) Both models' performance decreased as in-degree decreased, further indicating low in-degree worsens embedding learning performances, which is consistent with analysis results in section~\ref{exp:sparsity_analysis}. (3) Improvements from LR-GCN over CompGCN decrease as in-degree increases (except last in-degree scopes), revealing that LR-GCN is more sensitive to high-sparsity entities. We explain this phenomenon because high-sparsity entities can accrue more information gain from high-order graph structures. (4) Obvious improvements also occur at the maximal in-degree scopes on two datasets. We hypothesize that the great performance gains originate from the reasoning capacity distilled from MLN-RL, which benefits from the rich logical patterns that are inherent in dense graph structures. Future research will delve deeper into this hypothesis further.

\subsection{Comparison with K-stack Method} \label{exp:k-stack_comparison}
In this study, we evaluate the effectiveness of a proposed method that uses reasoning paths in RL to guide the exploration and exploitation of higher-order graph structural information. Our comparison involves LR-GCN and CompGCN with stacked graph convolutional layers on WD-singer and NELL23K datasets. The hyper-parameters used in CompGCN remain constant except for the number of layers K. Considering that we uniformly set the maximize reasoning steps implemented in MultihopKG~\cite{lin2018multi} as 3, the horizon of the GNN-based predictor has been broadened, allowing it to selectively see its 3-hop neighbors. Therefore we experiment with K=1,2,3 for comparative analysis. Our results in Table~\ref{tab:k_stack} demonstrate that CompGCN's performance cannot be improved by simply stacking graph convolution layers. This validates our statement about the over-squashing problem. Additionally, our proposed LR-GCN significantly outperforms CompGCN(K=2) and CompGCN(K=3), indicating that the high-order graph structure information obtained from our method is more useful for sparse KGC.

\begin{table}[t]
    \centering
    \caption{Performance comparison of LR-GCN with CompGCN stacked with $K=1,2,3$ graph convolutional layers on WD-singer and NELL23K.}
    % \resizebox{\linewidth}{!}{
    \begin{tabular}{cccccc}
    \toprule
    && \multicolumn{2}{c}{WD-singer} & \multicolumn{2}{c}{NELL23K}\\
    \cmidrule(lr){3-4}\cmidrule(lr){5-6}
    && Hits@10 & MRR & Hits@10 & MRR\\
    \midrule
    \multicolumn{2}{l}{CompGCN(K=1)} & 47.62 & 37.65 & 37.57 & 21.60\\
    \multicolumn{2}{l}{CompGCN(K=2)} & 47.98 & 36.55 & 37.34 & 21.13\\
    \multicolumn{2}{l}{CompGCN(K=3)} & 47.80 & 34.22 & 37.55 & 21.01\\
    \multicolumn{2}{l}{LR-GCN} & \textbf{49.98} & \textbf{39.27} & \textbf{41.90} & \textbf{25.27}\\
    \bottomrule
    \end{tabular}
    % }
    \label{tab:k_stack}
\end{table}

\begin{table}[t]
    \centering
    \caption{Performances of MLN-RL compared with its base model DacKGR on FB15K-237\_10, WD-singer and NELL23K. Values are in percentage.}
    % \resizebox{\linewidth}{!}{
    \begin{tabular}{cccccc}
    \toprule
    && \multicolumn{2}{c}{DacKGR} & \multicolumn{2}{c}{MLN-RL}\\
    \cmidrule(lr){3-4}\cmidrule(lr){5-6}
    && Hits@1 & MRR & Hits@1 & MRR\\
    \midrule
    \multicolumn{2}{l}{FB15K-237\_10} & 10.21 & 13.91 & 10.33 & 13.97\\
    \multicolumn{2}{l}{WD-singer} & 26.49 & 32.66 & 28.05 & 33.65\\
    \multicolumn{2}{l}{NELL23K} & 13.00 & 18.99 & 13.02 & 19.05\\
    \bottomrule
    \end{tabular}
    % }
    \label{tab:mln_performance}
\end{table}

% \subsection{Analysis of Loss weights}
% In section~\ref{optimization_evaluation}, we introduce two hyperparameters $\beta$ and $\gamma$ to trade off among three training losses. That is the bigger $\beta$ setting indicates that LR-GCN relies more on long-range convolution, while the bigger $\gamma$ means that the model pays more attention to higher-order knowledge distillation. Here we provide the influence of $\beta$ and $\gamma$ for NELL23K and FB15K-237\_10 in Table \ref{tab:hyperparameter_influence} for demonstration.

\begin{figure*}[t]
	\centering
	\subfigure[Case 1: (\textit{Intelligent dance music, parent\_genre, ?})]{
		\begin{minipage}{14cm}%[b]%{0.2\textwidth}
			\includegraphics[width=\textwidth]{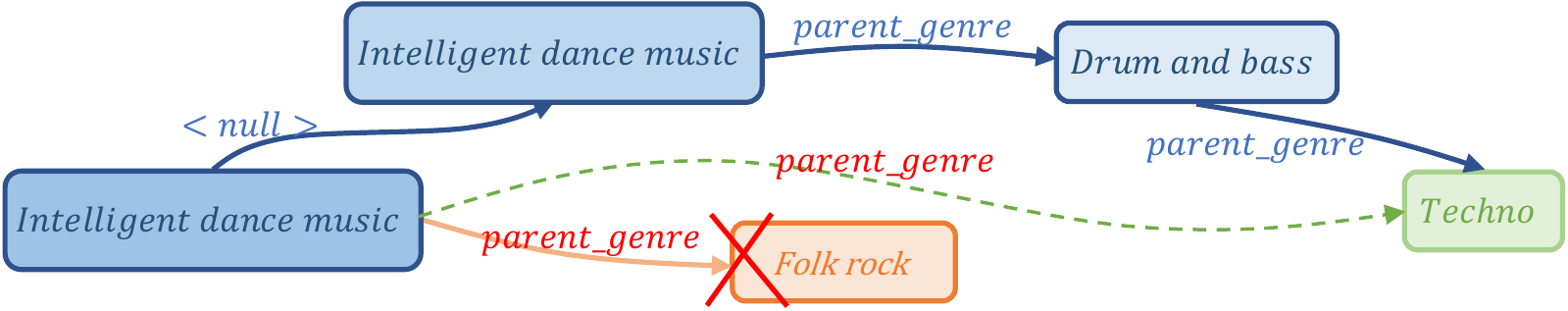} \\
			
		\end{minipage}
	}
    \subfigure[Case 2: (\textit{David Nutter, nominated\_for, ?})]{
		\begin{minipage}{14cm} %[b]%{0.2\textwidth} 
                        %{12cm or 0.2/textwidth} 控制图片大小，可以=textwidth
			% 插入子图片
                        \includegraphics[width=\textwidth]{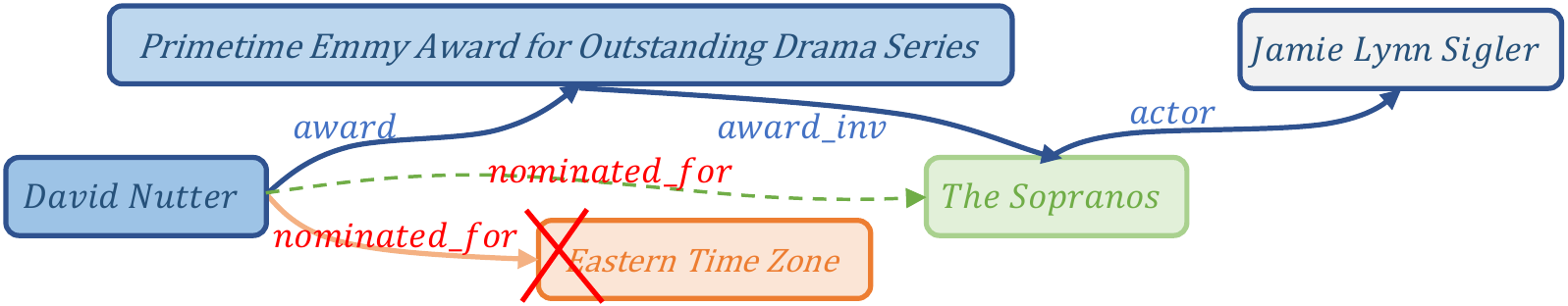} \\
			%\includegraphics[width=\textwidth]{fig/beforePAA.eps} 
                        % 可以在一个minipage里写多张图片，它们共用一个小标题\subfigure['sub_title']
		\end{minipage}
	}
    \caption{Blue edges denote the reasoning paths searched for corresponding queries. Orange and green nodes denote the predicted answers by the pre-trained GNN-based predictor before joint learning and the golden answers, respectively.}
    \label{fig:case_study}
\end{figure*}

\subsection{MLN-RL Performance} \label{mln-rl_performance}
We also propose a novel RL-based method MLN-RL to primarily enable existing RL-based methods to output probabilities to facilitate knowledge distillation for GNN-based models. Besides, the effect of the reasoning path distiller heavily impacts the final performance of the GNN-based predictor. Therefore, we train the MLN-RL module separately to observe its effect compared with DacKGR~\cite{lv2020dynamic}. We present DacKGR and MLN-RL results on FB15K-237\_10, WD-singer, and NELL23K in Table~\ref{tab:mln_performance}. The results demonstrate that MLN-RL outperforms DacKGR, especially on WD-singer, which verifies the feasibility and security of MLN-RL.

% For more analysis experiments please refer to Appendix~\ref{app:more_analysis_exp}

% \input{Tables/hyperparameter_influence.tex}

% \subsection{Case Study}
% To provide an intuitive impression of the rationality and validity of explicit modeling for reasoning paths, we take two instances for illustration, shown in Tab.\ref{tab:case_studies}, in the first case, LR-GCN corrects false prediction with the help of the rational path, and then LR-GCN compress

\subsection{Case Study} \label{exp:case_study}
We visualize some exemplars learned by LR-GCN on the FB15K-237\_10 dataset in Fig.~\ref{fig:case_study} to further explain our motivations. The first instance inquires about the \textit{parent genre} of \textit{Intelligent dance music}. Prior to joint learning, the CompGCN model predicts an incorrect response \textit{Folk rock} but successfully predicts the correct answer with the aid of the positive reasoning path found by the MLN-RL module. This can be attributed to the establishment of a closer semantic connection between \textit{Intelligent dance music} and \textit{Techno}, as well as the possible injection of logic rules $parent\_genre\wedge parent\_genre\rightarrow parent\_genre$ during training. 
% The earnings come from two folds: on the one hand, a direct connection between \textit{Intelligent dance music} and \textit{Techo} increases their relevance; and the possible injection of logic rules $parent\_genre\wedge parent\_genre\rightarrow parent\_genre$ during the training phase might also help the prediction. 
The second instance demonstrates a negative reasoning path where the predicted answer is erroneous; however, the intermediate answer in the reasoning path aids in correcting the prediction during joint learning. Similar analyses can also be conducted upon this demonstration.

\section{Related Work}
% \subsection{Knowledge Graph Embedding}
\noindent\textbf{Knowledge Graph Completion}
Knowledge Graph Embedding (KGE) serves the most common method for Knowledge Graph Completion, which aims to learn dense distributed representation for entities and relations in KGs. According to the design criteria of the scoring function, KGE approaches can be classified into three main families: translation-based models, semantic matching-based, and deep learning models \cite{rossi2021knowledge}, represented by TransE \cite{bordes2013translating}, DistMult \cite{yang2015embedding}, and ConvE \cite{dettmers2018convolutional}, respectively. On top of the above KGE models, GNN-based KGC models encode graph structure first with Graph Neural Network (GNN) \cite{Kipf2016SemiSupervisedCW} and then score triples using mentioned KGE methods, represented by R-GCN~\cite{schlichtkrull2018modeling}, CompGCN~\cite{Vashishth2019CompositionbasedMG}, and SE-GNN~\cite{li2022does}.
% This strategy further improves the performance of KGE methods by virtue of the rich structural information contained in the encoded embeddings by GNN models. 
Reinforcement Learning (RL) can also be applied to multi-hop Knowledge Graph Completion \cite{lin2018multi,wan2021reasoning}, where the agent is trained to find the reasoning path for explaining the conducted query. The RL-based methods often lag behind KGE models, but provide better explainability.
However, previous KGC models assume the KGs are dense enough to learn rich-semantic embeddings, which is not common in real-world scenarios \cite{chen2022explainable}. On the contrary, KGs are often sparse due to the limitations of corpus and techniques. Obvious performance degradation can be observed for previous KGC models due to the growing sparsity of KGs.

\noindent\textbf{Sparse Knowledge Graph Completion}
To solve the problem of sparse KGC, \cite{lv2020dynamic} propose DacKGR, a Reinforcement Learning (RL) based method, by extending the action space using a pre-trained KGE model. The improvements of DacKGR are remarkable compared with its base method MultiHopKG~\cite{lin2018multi} but still lag behind KGE models. \cite{he2022vem} propose to apply textual information like entity names and descriptions to supplement insufficient features, but not all KGs contain textual information. \cite{chen2022explainable} propose HoGRN to relieve high sparsity from different views, which introduces weight-free attention and learns high-quality relation embeddings in the Graph Neural Network framework. \cite{tan2023kracl} also propose to aggregate messages with the attention mechanism and utilize contrastive learning to enrich entity semantics. Obviously, these two works are confined to exploring the efficient use of 1-hop neighbors information without considering introducing higher-order graph structure features to enrich semantics of entities.

% \subsection{KGC based on Markov Logic Network}
\noindent\textbf{KGC based on Markov Logic Network} Markov Logic Network (MLN) \cite{richardson2006markov} defines the joint distribution of the observed and the hidden variables in undirected graphical models employing the first-order logic. \cite{qu2019probabilistic} and \cite{zhang2020efficient} apply MLN in KGs to distill logic rules into embedding learning via data augmentation. 
However, MLN has faced criticism due to its high computational complexity, specifically in inducing formulas and grounding paths. Moreover, \cite{qu2019probabilistic,zhang2020efficient} also did not consider graph densification and 
explicitly integrating high-order graph structure within graph convolutional layers.
This study enhances the computational efficiency of MLN by directly inducing rules from reasoning paths generated by Reinforcement Learning-based models and enables dynamic updating of the rule base. Detailed information on this approach can be found in Section~\ref{Reasoning Path Extractor}.

% However, MLN is often criticized for computational complexity, including inducing formulas and groundings. In contrast to MLN, we improve the computational efficiency of MLN by sacrificing accuracy by directly inducing rules from reasoning paths produced by RL-based models. More details can be found in section \ref{Reasoning Path Extractor}.

\section{Conclusions}
In this paper, we focus on the task of sparse Knowledge Graph Completion and attribute this challenge to the inadequate information from direct neighbors. This paper proposes a novel GNN-based framework, LR-GCN, that addresses this challenge by leveraging high-order graph structure information to enrich entity semantics. LR-GCN comprises two components: a base GNN-based model and an RL-based reasoning path distiller MLN-RL. MLN-RL learns to explore meaningful reasoning paths. And then two different strategies are proposed to fully exploit long-range dependency knowledge contained in reasoning paths, which learn complementary knowledge. Experimental results support the effectiveness of LR-GCN in addressing sparse KGC. Future work may involve incorporating additional features such as text information and improving the reasoning abilities of GNN-based methods.

% \paragraph*{Supplemental Material Statement:} Full proofs of Theorem 1, details of datasets, detailed configurations, and source code are attached with the submission in EasyChair and, if accepted, will be published on arXiv in an extended version of the paper.


\bibliography{main}

\begin{thebibliography}{34}
\providecommand{\natexlab}[1]{#1}
\providecommand{\url}[1]{\texttt{#1}}
\expandafter\ifx\csname urlstyle\endcsname\relax
  \providecommand{\doi}[1]{doi: #1}\else
  \providecommand{\doi}{doi: \begingroup \urlstyle{rm}\Url}\fi

\bibitem[Bala{\v{z}}evi{\'c} et~al.(2019)Bala{\v{z}}evi{\'c}, Allen, and
  Hospedales]{balavzevic2019tucker}
Bala{\v{z}}evi{\'c}, I., Allen, C., and Hospedales, T.
\newblock Tucker: Tensor factorization for knowledge graph completion.
\newblock In \emph{Proceedings of the 2019 Conference on Empirical Methods in
  Natural Language Processing and the 9th International Joint Conference on
  Natural Language Processing (EMNLP-IJCNLP)}, pp.\  5185--5194, 2019.

\bibitem[Bishop \& Nasrabadi(2006)Bishop and Nasrabadi]{bishop2006pattern}
Bishop, C.~M. and Nasrabadi, N.~M.
\newblock \emph{Pattern recognition and machine learning}, volume~4.
\newblock Springer, 2006.

\bibitem[Bordes et~al.(2013)Bordes, Usunier, Garcia-Duran, Weston, and
  Yakhnenko]{bordes2013translating}
Bordes, A., Usunier, N., Garcia-Duran, A., Weston, J., and Yakhnenko, O.
\newblock Translating embeddings for modeling multi-relational data.
\newblock \emph{Advances in neural information processing systems}, 26, 2013.

\bibitem[Cao et~al.(2022)Cao, Shi, Pan, Nie, Xiang, Hou, Li, He, and
  Zhang]{cao2022kqa}
Cao, S., Shi, J., Pan, L., Nie, L., Xiang, Y., Hou, L., Li, J., He, B., and
  Zhang, H.
\newblock Kqa pro: A dataset with explicit compositional programs for complex
  question answering over knowledge base.
\newblock In \emph{Proceedings of the 60th Annual Meeting of the Association
  for Computational Linguistics (Volume 1: Long Papers)}, pp.\  6101--6119,
  2022.

\bibitem[Chen et~al.(2022)Chen, Cao, Feng, He, and Zhang]{chen2022explainable}
Chen, W., Cao, Y., Feng, F., He, X., and Zhang, Y.
\newblock Explainable sparse knowledge graph completion via high-order graph
  reasoning network.
\newblock \emph{arXiv preprint arXiv:2207.07503}, 2022.

\bibitem[Dettmers et~al.(2018)Dettmers, Minervini, Stenetorp, and
  Riedel]{dettmers2018convolutional}
Dettmers, T., Minervini, P., Stenetorp, P., and Riedel, S.
\newblock Convolutional 2d knowledge graph embeddings.
\newblock In \emph{Thirty-second AAAI conference on artificial intelligence},
  2018.

\bibitem[Fei et~al.(2022)Fei, Zhou, Gui, Zhang, and Huang]{fei2022lfkqg}
Fei, Z., Zhou, X., Gui, T., Zhang, Q., and Huang, X.-J.
\newblock Lfkqg: A controlled generation framework with local fine-tuning for
  question generation over knowledge bases.
\newblock In \emph{Proceedings of the 29th International Conference on
  Computational Linguistics}, pp.\  6575--6585, 2022.

\bibitem[Galkin et~al.(2022)Galkin, Zhu, Ren, and Tang]{galkin2022inductive}
Galkin, M., Zhu, Z., Ren, H., and Tang, J.
\newblock Inductive logical query answering in knowledge graphs.
\newblock \emph{Advances in Neural Information Processing Systems},
  35:\penalty0 15230--15243, 2022.

\bibitem[He et~al.(2022)He, Jiang, Zheng, Zhu, Zhang, Liu, Zhao, and
  Qin]{he2022vem}
He, T., Jiang, T., Zheng, Z., Zhu, H., Zhang, J., Liu, M., Zhao, S., and Qin,
  B.
\newblock Vem$^{2}$l: A plug-and-play framework for fusing text and structure
  knowledge on sparse knowledge graph completion.
\newblock \emph{arXiv preprint arXiv:2207.01528}, 2022.

\bibitem[Jin et~al.(2022)Jin, Gong, Wang, Yu, He, Huang, and
  Wang]{jin2022graph}
Jin, D., Gong, Y., Wang, Z., Yu, Z., He, D., Huang, Y., and Wang, W.
\newblock Graph neural network for higher-order dependency networks.
\newblock In \emph{Proceedings of the ACM Web Conference 2022}, pp.\
  1622--1630, 2022.

\bibitem[Kipf \& Welling(2016)Kipf and Welling]{Kipf2016SemiSupervisedCW}
Kipf, T. and Welling, M.
\newblock Semi-supervised classification with graph convolutional networks.
\newblock \emph{ArXiv}, abs/1609.02907, 2016.

\bibitem[Li et~al.(2022{\natexlab{a}})Li, Li, Zhang, Li, Wei, Cui, and
  Wang]{li2022c3kg}
Li, D., Li, Y., Zhang, J., Li, K., Wei, C., Cui, J., and Wang, B.
\newblock C3kg: A chinese commonsense conversation knowledge graph.
\newblock In \emph{Findings of the Association for Computational Linguistics:
  ACL 2022}, pp.\  1369--1383, 2022{\natexlab{a}}.

\bibitem[Li et~al.(2022{\natexlab{b}})Li, Cao, Zhu, Bi, Fang, Liu, and
  Li]{li2022does}
Li, R., Cao, Y., Zhu, Q., Bi, G., Fang, F., Liu, Y., and Li, Q.
\newblock How does knowledge graph embedding extrapolate to unseen data: a
  semantic evidence view.
\newblock In \emph{Proceedings of the AAAI Conference on Artificial
  Intelligence}, 2022{\natexlab{b}}.

\bibitem[Lin et~al.(2018)Lin, Socher, and Xiong]{lin2018multi}
Lin, X.~V., Socher, R., and Xiong, C.
\newblock Multi-hop knowledge graph reasoning with reward shaping.
\newblock In \emph{Proceedings of the 2018 Conference on Empirical Methods in
  Natural Language Processing}, pp.\  3243--3253, 2018.

\bibitem[Lv et~al.(2020)Lv, Han, Hou, Li, Liu, Zhang, Zhang, Kong, and
  Wu]{lv2020dynamic}
Lv, X., Han, X., Hou, L., Li, J., Liu, Z., Zhang, W., Zhang, Y., Kong, H., and
  Wu, S.
\newblock Dynamic anticipation and completion for multi-hop reasoning over
  sparse knowledge graph.
\newblock In \emph{Proceedings of the 2020 Conference on Empirical Methods in
  Natural Language Processing (EMNLP)}, pp.\  5694--5703, 2020.

\bibitem[Qu \& Tang(2019)Qu and Tang]{qu2019probabilistic}
Qu, M. and Tang, J.
\newblock Probabilistic logic neural networks for reasoning.
\newblock \emph{Advances in neural information processing systems}, 32, 2019.

\bibitem[Richardson \& Domingos(2006)Richardson and
  Domingos]{richardson2006markov}
Richardson, M. and Domingos, P.
\newblock Markov logic networks.
\newblock \emph{Machine learning}, 62\penalty0 (1):\penalty0 107--136, 2006.

\bibitem[Rossi et~al.(2021)Rossi, Barbosa, Firmani, Matinata, and
  Merialdo]{rossi2021knowledge}
Rossi, A., Barbosa, D., Firmani, D., Matinata, A., and Merialdo, P.
\newblock Knowledge graph embedding for link prediction: A comparative
  analysis.
\newblock \emph{ACM Transactions on Knowledge Discovery from Data (TKDD)},
  15\penalty0 (2):\penalty0 1--49, 2021.

\bibitem[Schlichtkrull et~al.(2018)Schlichtkrull, Kipf, Bloem, Van Den~Berg,
  Titov, and Welling]{schlichtkrull2018modeling}
Schlichtkrull, M., Kipf, T.~N., Bloem, P., Van Den~Berg, R., Titov, I., and
  Welling, M.
\newblock Modeling relational data with graph convolutional networks.
\newblock In \emph{European semantic web conference}, pp.\  593--607. Springer,
  2018.

\bibitem[Shang et~al.(2019)Shang, Tang, Huang, Bi, He, and Zhou]{shang2019end}
Shang, C., Tang, Y., Huang, J., Bi, J., He, X., and Zhou, B.
\newblock End-to-end structure-aware convolutional networks for knowledge base
  completion.
\newblock In \emph{Proceedings of the AAAI Conference on Artificial
  Intelligence}, 2019.

\bibitem[Sui et~al.(2023)Sui, Zeng, Chen, Liu, and Zhao]{sui2023joint}
Sui, D., Zeng, X., Chen, Y., Liu, K., and Zhao, J.
\newblock Joint entity and relation extraction with set prediction networks.
\newblock \emph{IEEE Transactions on Neural Networks and Learning Systems},
  2023.

\bibitem[Sun et~al.(2018)Sun, Deng, Nie, and Tang]{Sun2018RotatEKG}
Sun, Z., Deng, Z., Nie, J.-Y., and Tang, J.
\newblock Rotate: Knowledge graph embedding by relational rotation in complex
  space.
\newblock \emph{ArXiv}, abs/1902.10197, 2018.

\bibitem[Sun et~al.(2020)Sun, Vashishth, Sanyal, Talukdar, and Yang]{sun2020re}
Sun, Z., Vashishth, S., Sanyal, S., Talukdar, P., and Yang, Y.
\newblock A re-evaluation of knowledge graph completion methods.
\newblock In \emph{Proceedings of the 58th Annual Meeting of the Association
  for Computational Linguistics}, pp.\  5516--5522, 2020.

\bibitem[Tan et~al.(2023)Tan, Chen, Feng, Zhang, Zheng, Li, and
  Luo]{tan2023kracl}
Tan, Z., Chen, Z., Feng, S., Zhang, Q., Zheng, Q., Li, J., and Luo, M.
\newblock Kracl: contrastive learning with graph context modeling for sparse
  knowledge graph completion.
\newblock In \emph{Proceedings of the ACM Web Conference 2023}, pp.\
  2548--2559, 2023.

\bibitem[Topping et~al.(2021)Topping, Giovanni, Chamberlain, Dong, and
  Bronstein]{Topping2021UnderstandingOA}
Topping, J., Giovanni, F.~D., Chamberlain, B.~P., Dong, X., and Bronstein,
  M.~M.
\newblock Understanding over-squashing and bottlenecks on graphs via curvature.
\newblock \emph{ArXiv}, abs/2111.14522, 2021.

\bibitem[Trouillon et~al.(2016)Trouillon, Welbl, Riedel, Gaussier, and
  Bouchard]{trouillon2016complex}
Trouillon, T., Welbl, J., Riedel, S., Gaussier, {\'E}., and Bouchard, G.
\newblock Complex embeddings for simple link prediction.
\newblock In \emph{International conference on machine learning}, pp.\
  2071--2080. PMLR, 2016.

\bibitem[Vashishth et~al.(2019)Vashishth, Sanyal, Nitin, and
  Talukdar]{Vashishth2019CompositionbasedMG}
Vashishth, S., Sanyal, S., Nitin, V., and Talukdar, P.~P.
\newblock Composition-based multi-relational graph convolutional networks.
\newblock \emph{ArXiv}, abs/1911.03082, 2019.

\bibitem[Wan et~al.(2021)Wan, Pan, Gong, Zhou, and Haffari]{wan2021reasoning}
Wan, G., Pan, S., Gong, C., Zhou, C., and Haffari, G.
\newblock Reasoning like human: Hierarchical reinforcement learning for
  knowledge graph reasoning.
\newblock In \emph{Proceedings of the twenty-ninth international conference on
  international joint conferences on artificial intelligence}, pp.\
  1926--1932, 2021.

\bibitem[Xu et~al.(2023)Xu, Zhu, Wang, and Zhang]{Xu2023HowTU}
Xu, X., Zhu, Y., Wang, X., and Zhang, N.
\newblock How to unleash the power of large language models for few-shot
  relation extraction?
\newblock \emph{ArXiv}, abs/2305.01555, 2023.

\bibitem[Yang et~al.(2015)Yang, Yih, He, Gao, and Deng]{yang2015embedding}
Yang, B., Yih, S. W.-t., He, X., Gao, J., and Deng, L.
\newblock Embedding entities and relations for learning and inference in
  knowledge bases.
\newblock In \emph{Proceedings of the International Conference on Learning
  Representations (ICLR) 2015}, 2015.

\bibitem[Yang et~al.(2018)Yang, Liu, Zheng, and Han]{yang2018node}
Yang, C., Liu, M., Zheng, V.~W., and Han, J.
\newblock Node, motif and subgraph: Leveraging network functional blocks
  through structural convolution.
\newblock In \emph{2018 IEEE/ACM International Conference on Advances in Social
  Networks Analysis and Mining (ASONAM)}, pp.\  47--52. IEEE, 2018.

\bibitem[Zhang \& Yao(2022)Zhang and Yao]{zhang2022knowledge}
Zhang, Y. and Yao, Q.
\newblock Knowledge graph reasoning with relational digraph.
\newblock In \emph{Proceedings of the ACM Web Conference 2022}, pp.\  912--924,
  2022.

\bibitem[Zhang et~al.(2020)Zhang, Chen, Yang, Ramamurthy, Li, Qi, and
  Song]{zhang2020efficient}
Zhang, Y., Chen, X., Yang, Y., Ramamurthy, A., Li, B., Qi, Y., and Song, L.
\newblock Efficient probabilistic logic reasoning with graph neural networks.
\newblock \emph{arXiv preprint arXiv:2001.11850}, 2020.

\bibitem[Zhu et~al.(2021)Zhu, Zhang, Xhonneux, and Tang]{zhu2021neural}
Zhu, Z., Zhang, Z., Xhonneux, L.-P., and Tang, J.
\newblock Neural bellman-ford networks: A general graph neural network
  framework for link prediction.
\newblock \emph{Advances in Neural Information Processing Systems},
  34:\penalty0 29476--29490, 2021.

\end{thebibliography}
\bibliographystyle{icml2021}

\appendix
\appendix

\newpage

% \input{Tables/dataset_statistics.tex}

% \section{Proofs}
\section{Proof for the Optimization Objective in the E-Step} \label{app:proof}
% \begin{theorem}
\noindent\textbf{Theorem 1.}
\textit{
Optimize $E_{p_\phi(t_c)}[w_{rule[g]} w_\theta(e_n|g,\mathcal{Q})]$ by gradient descent approximates to optimize $E_{p_\phi(t_c)}[\log p_\phi(t_c)[w_{rule[g]} w_\theta(e_n|g,\mathcal{Q})]]$:}
% \end{theorem}
\begin{proof}
\begin{equation}
    \begin{aligned}
        &\nabla_\phi E_{p_\phi(t_c)}[w_{rule[g]} w_\theta(e_n|g,\mathcal{Q})]\\
        =&\int \nabla_\phi p_\phi(t_c)[w_{rule[g]}w_\theta(e_n|g,\mathcal{Q})]\\
        =&\int p_\phi(t_c) \nabla_\phi \log p_\phi(t_c)[w_{rule[g]}w_\theta(e_n|g,\mathcal{Q})]\\
        =&E_{p_\phi(t_c)}[\nabla_\phi \log p_\phi(t_c)[w_{rule[g]}w_\theta(e_n|g,\mathcal{Q})]]\\
        \approx&\nabla_\phi \log p_\phi(t_c)[w_{rule[g]}w_\theta(e_n|g,\mathcal{Q})]
    \end{aligned}
\end{equation}
\end{proof}

\section{Construction Details of Datasets} \label{dataset_construction_details}
Similar to \cite{lv2020dynamic}, we also extract different proportions of triples from FB15K-237 to construct sparse KGs. Instead of randomly extracting triples as \cite{lv2020dynamic} do, we control that the number of entities and relationships remains unchanged in sub-datasets after extraction processes, i.e. we ensure that each entity or relationship participates in at least one triple. We construct sub-datasets in this way for two reasons: (1) We are able to control the sparsity of the extracted subgraphs precisely since the average in-degree is only determined by the number of triples if the number of entities in the subgraphs remains constant. (2) Retaining all entities and relations ensures that the validation and test sets remain unchanged, and we believe that it is fairer to compare model performances on datasets with different sparsity.

\begin{table}[htbp]
    \centering
    \caption{Part of best hyperparameter settings and running time of LR-GCN for different datasets. ``TT" means ``Training time".}
    % \resizebox{\linewidth}{!}{
    \begin{tabular}{cccccc}
    \toprule
    && $\beta$ & $\gamma$ & $\lambda$ & TT\\
    \midrule
    \multicolumn{2}{l}{FB15K-237\_10} & 1.0 & 5.0 & 0.0 & 4h\\
    \multicolumn{2}{l}{FB15K-237\_20} & 2.0 & 5.0 & 0.0001 & 13.5h\\
    \multicolumn{2}{l}{FB15K-237\_30} & 1.0 & 1.0 & 0.001 & 12.5h\\
    \multicolumn{2}{l}{FB15K-237\_60} & 2.0 & 0.1 & 0.0 & 15.5h\\
    \multicolumn{2}{l}{WD-singer} & 1.0 & 1.0 & 0.001 & 9h\\
    \multicolumn{2}{l}{NELL23K} & 0.5 & 2.0 & 0.0 & 2h\\
    \bottomrule
    \end{tabular}
    % }
    \label{tab:hyperparameters}
\end{table}

\section{Hyperparameter Settings} \label{app:hyperparamter_setting}
% For experiments, we choose between MultihopKG \cite{lin2018multi} and DacKGR \cite{lv2020dynamic} to implement the RL-based model. Intuitively, DacKGR is further ideal since its performance improvements better on sparse KGs against MultihopKG. However, we find that paths searched by DacKGR are less interpretable (even if prediction answers are correct), which may lead to weak improvements in some datasets. We 
Our model is implemented using PyTorch, running with Tesla\_V100-SXM2-16gb. The running times for each dataset are presented in Table~\ref{tab:hyperparameters}.
The training batch size is uniformly set to 128 for all experiments. Loss weights $\beta$ and $\gamma$ mentioned in Eq.\eqref{eq:gnn_loss} are tuned in $\{0.1, 0.5, 1.0, 2.0, 3.0, 4.0, 5.0\}$, and $\lambda$ in Eq.\eqref{eq:elbo_loss} are tuned in $\{0.0, 0.0001, 0.001, 0.01\}$. Best loss weight hyperparameters are presented in Table~\ref{tab:hyperparameters}.

% \input{Tables/hyperparameter_influence.tex}

% \section{More Analysis Results}     \label{app:more_analysis_exp}
% \subsection{Analysis of Loss weights}
% In section~\ref{optimization_evaluation}, we introduce two hyperparameters $\beta$ and $\gamma$ to trade off among three training losses. That is the bigger $\beta$ setting indicates that LR-GCN relies more on long-range convolution, while the bigger $\gamma$ means that the model pays more attention to higher-order knowledge distillation. Here we provide the influence of $\beta$ and $\gamma$ for NELL23K and FB15K-237\_10 in Table \ref{tab:hyperparameter_influence_nell} and \ref{tab:hyperparameter_influence_fb} respectively for demonstration.

\end{document}